\documentclass[preprint,12pt]{amsart}

\usepackage[foot]{amsaddr}

\usepackage[margin=1in]{geometry}              
\usepackage{amsmath} 
\usepackage{bbm}
\usepackage{xcolor}

\usepackage{subcaption}
\usepackage{tikz}

\usepackage{amsfonts}              
\usepackage{amsthm}
\usepackage{enumerate}

\usepackage[linesnumbered,ruled,vlined]{algorithm2e}

\SetKwInput{KwInput}{Input}                
\SetKwInput{KwOutput}{Output}              

\numberwithin{equation}{section}

\usepackage{graphicx}
\graphicspath{ {./images/} }

\newtheorem{thm}{Theorem}[section]

\newtheorem{Rem}[thm]{Remark}

\def\XXint#1#2#3{{\setbox0=\hbox{$#1{#2#3}{\int}$}
     \vcenter{\hbox{$#2#3$}}\kern-.5\wd0}}

\title[ A physics-informed neural network framework]{ A physics-informed neural network framework for modeling obstacle-related equations}
\author{Hamid El Bahja}
\address{Hamid EL Bahja (Corresponding author), African Institute for Mathematical Sciences, Research and Innovation Centre, Kigali, Rwanda}
\email{helbahja@aims.ac.za}
\author{Jan C. Hauffen}
\address{Jan Christian Hauffen, Technical University Berlin, Germany}
\email{j.hauffen@tu-berlin.de}
\author{Peter Jung}
\address{Peter Jung, Technical University Berlin and German Aerospace Center (DLR), Germany}
\email{peter.jung@tu-berlin.de}
\author{Bubacarr Bah}
\address{Bubacarr Bah, AIMS, Cape Town, South Africa}
\email{bubacarr@aims.ac.za}
\author{Issa Karambal}
\address{Issa Karambal,  Quantum leap Africa, Rwanda}
\email{ikarambal@quantumleapafrica.org}
\begin{document}

\begin{abstract}
Deep learning, a subset of machine learning, involves using neural networks with many layers to model and understand complex patterns in data. It has achieved remarkable success across various fields such as computer vision, natural language processing, and more recently, in solving partial differential equations (PDEs). Physics-informed neural networks (PINNs) are a novel approach that leverages deep learning to solve PDEs by incorporating physical laws directly into the neural network’s training process. This allows PINNs to handle scenarios with sparse and noisy data effectively. In this work, we extend the application of PINNs to solve obstacle-related PDEs, which are particularly challenging as they require numerical methods that can accurately approximate solutions constrained by obstacles. These obstacles can represent physical barriers or constraints in the solution space. We specifically focus on employing PINNs with hard boundary conditions. Hard boundary conditions ensure that the solution strictly adheres to the constraints imposed by the physical boundaries, rather than merely approximating them. This is achieved by explicitly encoding the boundary conditions into the neural network architecture, which guarantees that the solutions satisfy these conditions throughout the training process. By using PINNs with hard boundary conditions, we can effectively address the complexities posed by both regular and irregular obstacle configurations, ensuring that the solutions remain physically realistic and mathematically precise. We demonstrate the efficacy of the proposed PINNs in various scenarios involving both linear and nonlinear PDEs, with both regular and irregular obstacle constraints. \end{abstract}
 \maketitle
\section{Introduction}
The classical obstacle problem aims to describe the shape of an elastic membrane lying above an obstacle. Geometrically speaking, consider an elastic membrane that takes a specific shape under the application of a particular force. If an obstacle is close enough to the membrane, and it is in the direction of the applied force, the membrane will deform because of the obstacle. Obstacle problems have strong physical and engineering backgrounds. They emerge for instance, for the mathematical description of many physical systems including elastic-plastic torsion \cite{tor}, phase transition \cite{tran}, membrane-fluid interaction \cite{mem}, shallow ice sheets \cite{ice,ice2}, etc., with typical applications varying from tumor growth modeling \cite{chap}, to chemical vapor deposition \cite{fri}, crystal growth and solidification in materials \cite{mad}, semi-conductor design \cite{fri2}, and option pricing \cite{pha}.

Throughout the years, several numerical methods have been developed to solve different types of obstacle problems typically focusing on solving the variational inequality that appears from the first variation of the constrained problem. For instance, a finite difference scheme based on the variational inequality and the use of a multigrid algorithm to speed up computations was studied in \cite{hop}.  In \cite{bad}, the finite element formulation of the variational inequality is solved using the Schwarz domain decomposition method. The convergence of the Schwarz domain decomposition for nonlinear variational inequalities is established in \cite{tai,bad2}. Alternative approaches use penalty formulations where the obstacle problem can also be relaxed to an unconstrained problem with the addition of a penalty term. In particular, in \cite{l1,l2,l3} the authors proposed an efficient numerical scheme for solving some obstacle problems based on a reformulation of the obstacle in terms of $L^1$ and $L^2$-like penalties on the variational problem. However, as expected the relaxed problem, using an $L^1$-penalty is non-differentiable,  and the $L^2$-penalty  which is parametrized with a coefficient that depends on a parameter $\varepsilon$ requires that the parameter goes to infinity in order for the solution to be exact. Recently, in \cite{xu} the authors introduce a penalty method where a penalized weak formulation, in the sense of \cite{l1,l3}, is minimized by using a deep neural network. Nevertheless, this method is tailored to each numerical example since the penalty parameters are tuned manually, which requires the knowledge of the exact solution a priori in order to have good approximations. For more numerical approaches see \cite{zo,kor,maj} and references therein.

Despite its effectiveness, each of the previously mentioned methods has its limitations such as lack of convergence, non-differentiability, and domain discretization dependency. Also, in most cases, these methods must be specifically tailored to a given problem setup and cannot be easily adapted to build a general framework for seamlessly tackling problems involving various types of partial differential equations related to an obstacle constraint. To bypass these limitations, we are going to use a Neural Network (NN) framework based on the high computational power of NNs in numerical estimations \cite{rag,pan}, on stochastic optimization \cite{erm}, on automatic differentiation, and the recent improvements in parallelized hardware\cite{ber,gun}. In particular, we concentrate on PINNs \cite{rai,dis,mil}. This approach has been used to approximate the solution of Allen–Cahn, Schr\"odinger, Navier–Stokes equations \cite{rud,rai}, and high-dimensional stochastic PDEs  \cite{han}. Moreover, PINNs have shown effectiveness in various engineering applications. In fluid dynamics, PINNs have been used for modeling turbulent flows \cite{cai}, simulating multiphase flows \cite{wen}, and optimizing aerodynamic designs \cite{mao}. In solid mechanics, PINNs have been applied to structural analysis \cite{hag}, and material characterization \cite{ami}. These studies demonstrate the potential of PINNs to enhance engineering simulations and design processes. As stated in \cite{rai}, we can consider this method as a class of reinforcement learning \cite{lan}, in which the learning consists of incentive maximization or loss minimization instead of direct training on data. Should the network prediction not meet a governing equation, it will lead to an escalation in the cost, and therefore the learning will follow a path that should minimize this cost. So far, most of the cases considered in the previous references are related to problems where the latent solution of the PDE has no constraints and no inter-facial phenomena occur which is the case for obstacle problems.

In this work, we extend the application of PINNs to solve obstacle-related PDEs, which are particularly challenging as they require numerical methods that can accurately approximate solutions constrained by obstacles. These obstacles can represent physical barriers or constraints in the solution space. We specifically focus on employing PINNs with hard boundary constraints to solve the problem of computing data-driven solutions to the partial differential equations of the following general form
\begin{equation}
\begin{cases}
\mathcal{N}[u](x)=f(x)~~~~&\text{in}~\Omega,\\
u(x)=g(x)~~~~&\text{on}~\partial\Omega,\\
u(x)\geq \varphi(x)~~~~&\text{in}~\Omega,
\end{cases}
\end{equation}
where $\mathcal{N}$ is a differential operator, $u: \overline{\Omega}\longrightarrow\mathbb{R}$ is the latent solution, $\Omega$ is a bounded domain with Lipschitz regular boundary in $\mathbb{R}^{N}$, $\overline{\Omega}$ and $\partial\Omega$ are respectively the closure and the boundary of $\Omega$, $f$ and $g$ are fixed values functions, and $\varphi: \overline{\Omega}\longrightarrow\mathbb{R}^{N}$ is a given obstacle.

Unlike traditional methods, PINNs with hard boundary conditions encode physical constraints directly into the neural network, ensuring strict adherence to these constraints throughout the training process. This approach not only enhances the accuracy of the solutions but also improves the model's ability to handle complex and irregular obstacle configurations. The effectiveness of PINNs with hard boundary constraints has been demonstrated in various studies. For example, Lagaris et al \cite{lag1,lag2} illustrated that embedding boundary conditions into neural networks significantly improved the accuracy of the solutions for boundary value problems. Moreover, PINNs with hard constraints were used for many forward and inverse problems, see for instance \cite{lu,su} and references therein.  

In this regard, our specific contributions can be summarized as follows:
\begin{itemize}
    \item We present a PINNs framework that can be applied to various linear and nonlinear PDEs subject to regular and irregular obstacles.
    \item We show the effectiveness of PINNs throughout many numerical experiments in solving various types of obstacle problems.
\end{itemize}
The paper is organized as follows. In Section 2, we give a mathematical and geometrical description of obstacle problems, In Section 3, we present a detailed description of the proposed PINNs framework. In Section 4, we demonstrate the effectiveness of PINNs through the lens of two representative case studies, including linear PDEs with regular and irregular obstacles, and nonlinear PDEs with regular and irregular obstacles. Finally, we conclude the paper in Section 5.

\section{Mathematical overview}
In this section, we take $\mathcal{N}[u]=\Delta u$ and $f=0$. Therefore, (1.1) becomes a problem of minimization of the following energy functional
\begin{equation}
    F(u)=\int_{\Omega}|\nabla u|^{2}~dx
\end{equation}
among all functions $u$ satisfying $u\geq\varphi$ in $\Omega$, for a given obstacle $\varphi$. The Euler-Lagrange equation of such a minimization problem is 
\begin{equation*}
    \begin{cases}
        u\geq\varphi~~~&\text{in}~\Omega,\\
        \Delta u=0~~~&\text{in}~\{u>\varphi\},\\
        -\Delta u\geq0~~~&\text{in}~\Omega.
    \end{cases}
\end{equation*}
This means that the solution $u$ is above the obstacle $\varphi$, it is harmonic when it doesn't meet the obstacle, and it is superharmonic throughout.
\begin{figure}[ht]
    \centering
    \includegraphics[width=0.7\textwidth, height=9cm]{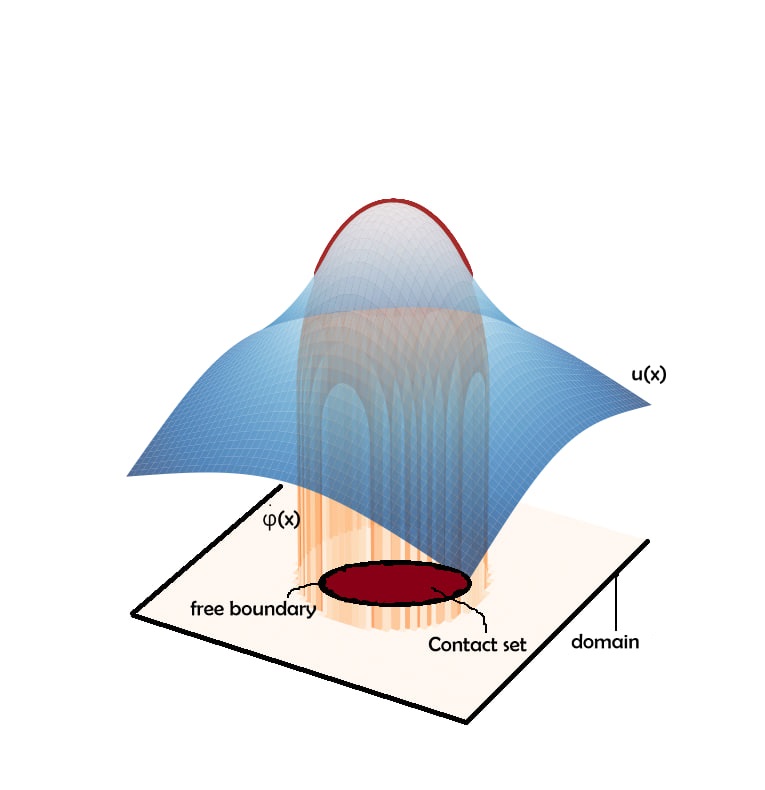}
    \caption{The contact set and the free boundary in the classical obstacle problem.}
\end{figure}

As illustrated in Figure 1, the domain $\Omega$ will be divided into two separate regions: one in which the solution $u$ is harmonic, and one in which the solution equals the obstacle. The last region is referred to as the contact set 
$\{u= \varphi\}\subset\Omega$. 
The area which separates these two regions is the free boundary. For example, in financial mathematics, this kind of problem emerges as a model for pricing American options \cite{pha,lau}. The function $\varphi$ stands for the option's payoff and the contact set $\{u = \varphi\}$ 
is the exercise region. In this context, the most significant unknown to figure out is the exercise region, i.e., one wants to investigate the two separate regions $\{x\in\Omega: u(x) = \varphi(x)\}$ in which we should apply the option, and $\{x\in\Omega: u(x) > \varphi(x)\}$ in which we would need to wait and not apply the option yet. The free boundary is the separating boundary between these two regions.  It is noteworthy that many studies guarantee the existence and the uniqueness of the solution to obstacle problems, see for example \cite{ham,bog} and references therein.

\section{Methodology}
In this study, we primarily focus on Dirichlet boundary conditions, extensively used in the literature on obstacle-related problems, as noted in \cite{rod,mem,bog} and references therein. To the best of our knowledge, there are no established examples of Neumann boundary conditions alone, making Dirichlet conditions the standard choice in the field. This choice ensures our method is comparable with existing studies. Additionally, we use the \textit{hard boundary method} to strictly impose Dirichlet boundary conditions by modifying the network architecture. This method ensures that the boundary conditions are satisfied exactly, which reduces the computational cost and simplifies implementation compared to traditional methods that incorporate boundary conditions through loss functions.

To this end, following the original work of Raissi et al. \cite{rai}, we assume that the solution \(u(x)\) of (1.1) can be approximated by using a feed-forward \(\alpha\)-layer neural network \(u(x;\theta)\), where \(\theta\) is a collection of all parameters in the network, such that
\begin{equation*}
 u(x;\theta)=\Sigma^{\alpha}\circ\Sigma^{\alpha-1}\circ\ldots\circ\Sigma^{1}(x),
\end{equation*}
where
\begin{equation*}
    \Sigma^{i}(z)=\sigma^{i}\left(W^{i}z+b^{i}\right),~~\text{for}~i=1,\ldots,\alpha.
\end{equation*}
In the above, the symbol \(\circ\) denotes the composition of functions, \(i\) is the layer number, \(x\) is the input to the network, \(\Sigma^{\alpha}\) is the output layer of the network, and \(W^{i}\) and \(b^{i}\) are respectively the weight matrices and bias vectors of layer \(i\), all collected in \(\theta=\{W^i,b^i\}_{i=1}^\alpha\). \(\sigma^{i}\) is the (point-wise) activation function for \(i=1,\ldots,\alpha-1\). In our numerical experiments below, for all the hidden layers, the hyperbolic-tangent function is used, which is a preferable activation function due to its smoothness and non-zero derivative.

To ensure that the neural network solution \(u(x;\theta)\) adheres to the boundary conditions of our obstacle problem, we implement the hard boundary method. Specifically, for a boundary condition \(u(x) = g(x)\) on the boundary \(\partial \Omega\), we define the network output as:
\begin{equation*}
    \hat{u}(x;\theta) = g(x) + \eta(x) \cdot u(x;\theta),
\end{equation*}
where \(\eta(x)\) is a smooth function that satisfies:
\begin{equation*}
    \begin{cases} 
       \eta(x) = 0 & \text{on}~\partial \Omega, \\
       \eta(x) >0 & \text{inside}~\Omega.
    \end{cases}
\end{equation*}
This construction ensures that the network output \(\hat{u}(x;\theta)\) equals the boundary condition \(g(x)\) on \(\partial \Omega\) while allowing the network \(u(x;\theta)\) to learn the solution structure inside the domain.

In summary, by using the hard boundary method, we embed the boundary conditions directly into the neural network's architecture, enhancing the capability of PINNs to solve boundary value problems accurately. Consequently, the parameters $\theta$ of $\hat{u}(x,\theta)$ can be learned by minimizing the following loss function
\begin{equation}
\mathcal{L}(\theta)=\frac{1}{N_r}\sum_{i=1}^{N_r}\left|H(\hat{u}(x_{r}^{i};\theta)-\varphi(x_{r}^{i}))\cdot R(x_{r}^{i};\theta)+\text{ReLu}(\varphi(x_{r}^{i})-\hat{u}(x_{r}^{i};\theta))\right|^{2},
\end{equation}
such that $R(x,\theta)$ is the PDE residual of (1.1) defined as
\begin{equation*}
    R(x;\theta)=\mathcal{N}[\hat{u}](x,\theta)-f(x),
\end{equation*}
$H$ is the Heaviside step function defined as
\begin{equation*}
    H(x)=\begin{cases}
    1~~~~&~\text{if}~x\geq0,\\
    0~~~~&~\text{otherwise},
    \end{cases}
\end{equation*}
$N_{r}$ denote the batch sizes for the training data $\{x_{r}^{i},f(x_{r}^{i})\}_{i=1}^{N_{r}}$, which can be randomly sampled at each iteration of a gradient descent algorithm. We can summarize the proposed PINNs method in Figure 2 and the following algorithm:
\begin{algorithm}[!ht]
\DontPrintSemicolon
  \KwInput{$\{x_r\}_{i=1}^{N_r},~tol>0$}
  Initialize to create the neural network in $\Omega$\\
  Predict the PINNs solution $\hat{u}( \cdot,\theta)$.\\
  Define the residual $R( \cdot,\theta)$.\\
  Define the $ReLu(\varphi( \cdot)-\hat{u}( \cdot,\theta))$ to penalize $\hat{u}( \cdot,\theta)$ that are under the obstacle $\varphi$.\\
  
  \While{$\mathcal{L}(\theta)>~tol$ }
    {
       Compute loss $\mathcal{L}(\theta)$.\\
       Train the loss $\mathcal{L}(\theta)$ by using Adam's optimizer.\\
       Updates weights and biases.\\}
   Fine-tuning using L-BFGS\\
  \KwOutput{$R(\hat{u}(x_r;\theta))\approx0,~\text{and}~\hat{u}( \cdot,\theta)\geq\varphi( \cdot)$}
\caption{PINNs method for obstacle problems}
\end{algorithm}
\begin{figure}[ht]
    \centering
    \includegraphics[
  width=12cm,
  height=5cm,
]{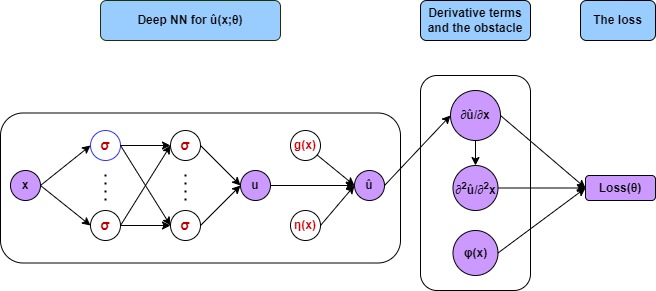}
    \caption{Neural Network Architecture for PINNs with Enforced Boundary Conditions for an Obstacle Problem.}
    \end{figure}

\begin{Rem}
Simultaneous training of the residual and obstacle loss by minimizing the following separated loss function:
\begin{equation}
\begin{split}
\mathcal{L}_\text{sep}(\theta) = 
& \frac{1}{N_r}\sum_{i=1}^{N_r} \left|H(\hat{u}(x_r^i; \theta) - \varphi(x_r^i)) \cdot R(x_r^i; \theta)\right|^2 \\
&+\frac{\lambda_{obs}}{N_r}\sum_{i=1}^{N_r} \left|\text{ReLu}(\varphi(x_r^i) - \hat{u}(x_r^i; \theta))\right|^2,
\end{split}
\end{equation}
where \(\lambda_{obs}\) denotes weight coefficients that are auto-tuned using back-propagated gradient statistics during training \cite{wan}, tends to converge very slowly in our case. For example, Figure 3 (orange line) shows that standard minimization algorithms take a long time to reduce the loss (3.2) to an acceptable level. 
\begin{figure}[ht]
    \centering
    \includegraphics[width=0.45\textwidth
]{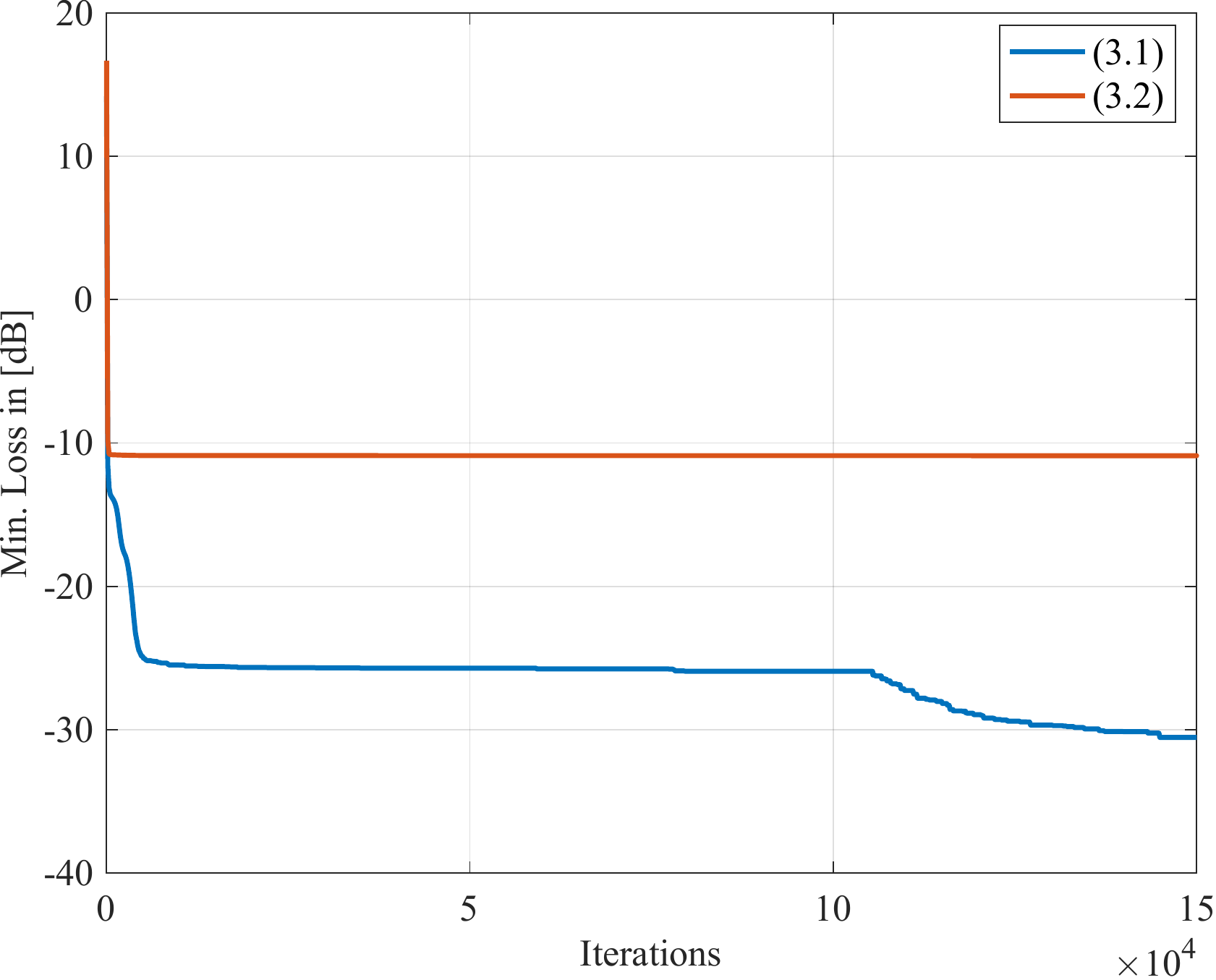}
    \caption{A comparison between the minimum loss over the number of iterations for the losses defined in (3.1) and (3.2).}
    \end{figure}
To address this issue, we proposed the reformulated loss function \(\mathcal{L}(\theta)\) as introduced in (3.1). Our approach selectively optimizes the network parameters based on the value of \(\hat{u}\) at each collocation point \(x_i\) relative to the obstacle \(\varphi\). The process is as follows:
\begin{itemize}

\item If \(\hat{u}(x_r^i; \theta) \geq \varphi(x_r^i)\), we focus on minimizing \(|R(x_r^i; \theta)|^2\) and ensure that \(|\text{ReLu}(\varphi(x_r^i) - \hat{u}(x_r^i; \theta))|^2 = 0\) since \(H(\hat{u}(x_r^i; \theta) - \varphi(x_r^i)) = 1\).
\item Conversely, if \(\hat{u}(x_r^i; \theta) < \varphi(x_r^i)\), we target minimizing \(|\text{ReLu}(\varphi(x_r^i) - \hat{u}(x_r^i; \theta))|^2\).
\end{itemize}
This implicit alternating optimization continues iteratively until the desired tolerance level is achieved. Figure 3 (blue line) shows that our method significantly accelerates convergence, demonstrating its effectiveness over the previous approach in (3.2).

\end{Rem}

\section{Numerical results}
 In this section, we apply the PINNs approach introduced above to various numerical examples. Throughout all case studies we will use fully-connected neural networks to approximate the latent functions representing PDE solutions and unknown boundaries where the used hyper-parameters for each obstacle are summarized in Table 1.
\begin{table}[th!]
\centering
\begin{tabular}{||c  c c c c||} 
 \hline
 Obstacle  & $N_r$ & Layers & Nodes & $tol$ \\ [0.5ex] 
 \hline\hline
 $\varphi_1$ & 5000 & 3 &24  & 3.8e-3 \\ 
 $\varphi_2$ & 5000 & 3  & 24 & 8.5e-3 \\
 $\varphi_3$ & 6300 & 6  &24  & 1.7e-3 \\
 $\varphi_4$ & 5000 & 3   &24  & 4e-5  \\
 $\varphi_5$ & 10000 & 3   & 24 & 2e-3 \\ 
 $\varphi_6$ & 15000 & 6  & 24  & 2.5e-3\\ [1ex] 
 \hline
\end{tabular}
\caption{Hyper-parameter settings employed throughout all numerical experiments presented in this work.}
\label{table:1}
\end{table}
The hyper-parameters listed in Table 1 were determined through extensive preliminary experiments to ensure optimal performance. In the following subsection, we will illustrate the process used to select these parameters using the first one-dimensional obstacle case as an example.
 \subsection{One-dimensional linear obstacle problems}
To check the performance of PINNs, we first restrict ourselves to studying a variety of one-dimensional regular obstacle problems which was previously considered in \cite{tran,sch}. Therefore, we have the following Poisson's equation
\begin{equation}
    -\frac{\partial^{2}u}{\partial^{2}x}=0,~~~\text{for}~x\in\Omega=(0,1),
\end{equation}
subject to Dirichlet boundary condition
\begin{equation}
    u(1)=u(0)=0,
\end{equation}
and to the following smooth one-dimensional obstacle 
 \begin{equation}
  \varphi_{1}(x)= \begin{cases}
    100x^2~~~~~&\text{for}~0\leq x\leq 0.25,\\
    100x(1-x)-12.5~~~~~&\text{for}~0.25\leq x\leq 0.75,\\
    100(1-x)^2~~~~~&\text{for}~0.75\leq x\leq 1,
      \end{cases}\end{equation}
such that $u\geq \varphi_{1}$   over $\Omega=(0,1)$. The correspondent  analytic solution of (4.1) under constraints (4.2) and (4.3) is
\begin{equation}u(x)=\begin{cases}
    (100-50\sqrt 2)x~~~~~&\text{for}~0\leq x\leq \frac{1}{2\sqrt 2},\\
    100x(1-x)-12.5~~~~~&\text{for}~\frac{1}{2\sqrt 2}\leq x\leq 1-\frac{1}{2\sqrt 2},\\
    (100-50\sqrt 2)(1-x)~~~~~&\text{for}~1-\frac{1}{2\sqrt 2}\leq x\leq 1.
    \end{cases}\end{equation}
Recall that Poisson's equation is one of the pivotal parts of electrostatics where the solution is the potential field caused by a given electric charge or mass density distribution. To this end, we will show that the deep neural network solution
\begin{equation}
\hat{u}(x;\theta)= x(1-x) u(x;\theta),
\end{equation}
specifically configured with $Nodes~=24$, $Layers~=3$, and collocation points $N_r=5000$ deduced from the preliminary experiments presented in Figure 4, approximates the latent solution $u(x)$ of (4.1) and satisfies the Dirichlet boundary condition and the obstacle constraint.
\begin{figure}[htbp]
    \centering
    \begin{subfigure}[b]{0.45\textwidth}
        \centering
        \includegraphics[width=\textwidth]{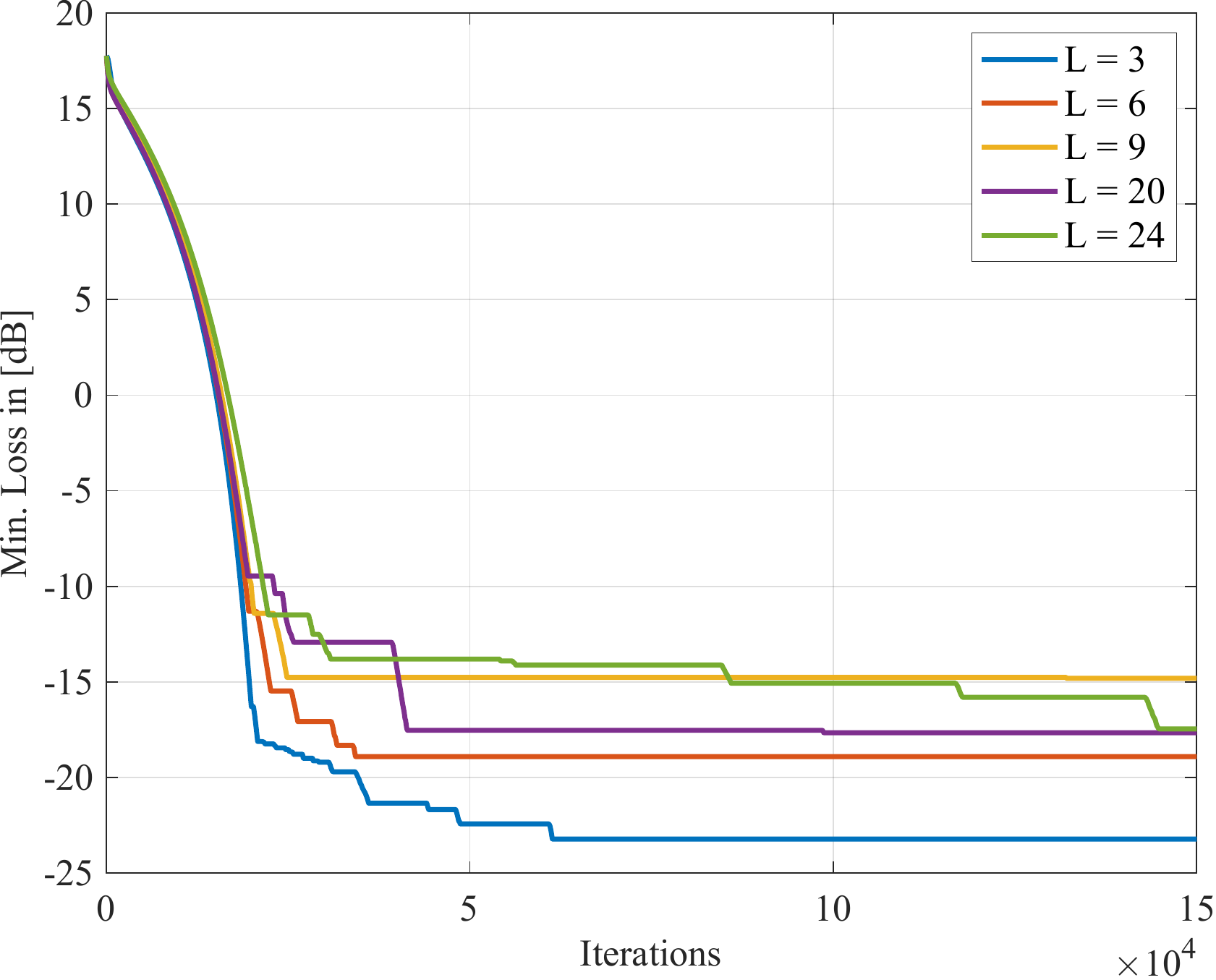}
        \caption*{(a)}
        \label{fig:subfigure1}
    \end{subfigure}
    \hfill
    \begin{subfigure}[b]{0.45\textwidth}
        \centering
        \includegraphics[width=\textwidth]{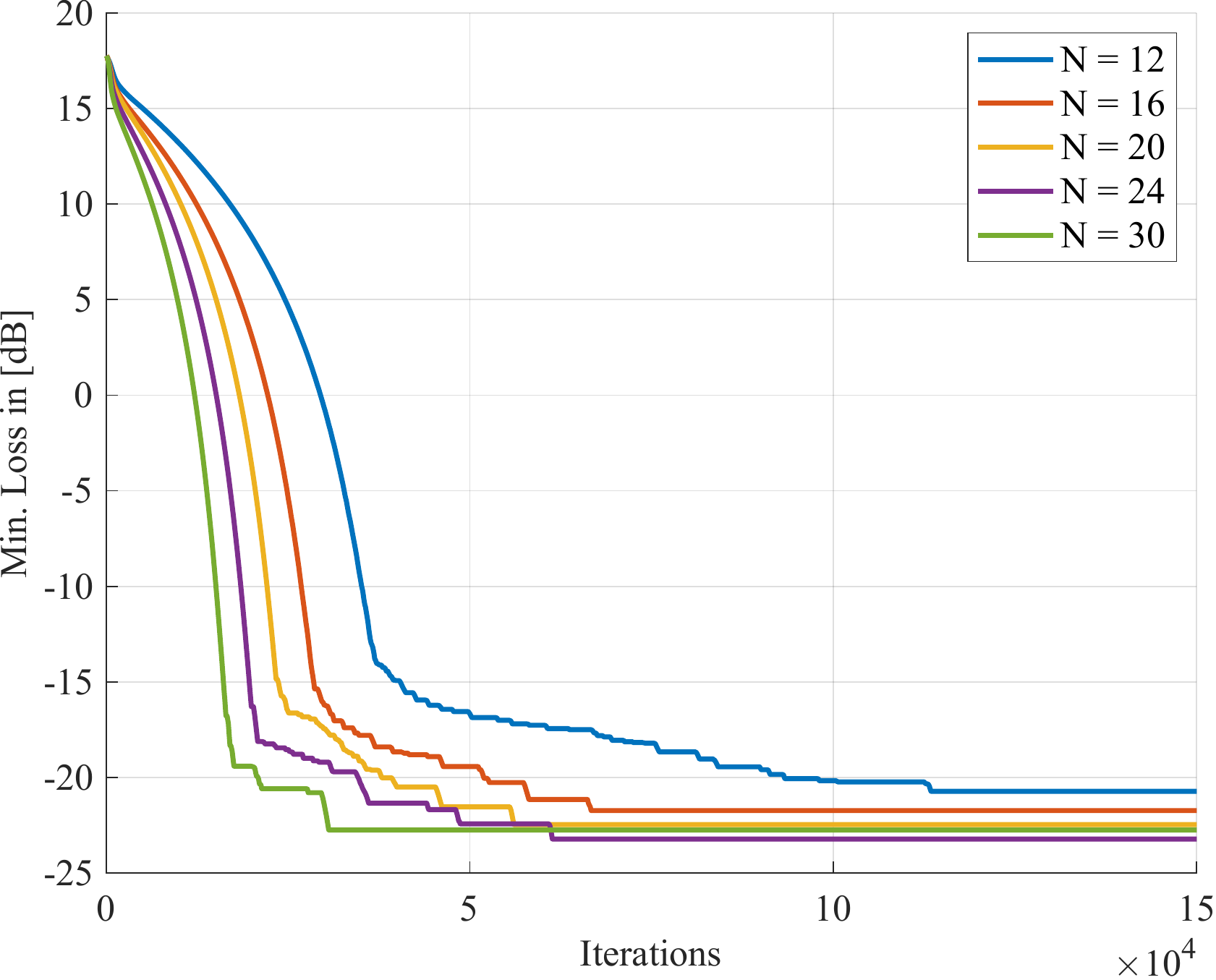}
        \caption*{(b)}
        \label{fig:subfigure2}
    \end{subfigure}
    \hfill
    \begin{subfigure}[b]{0.45\textwidth}
        \centering
        \includegraphics[width=\textwidth]{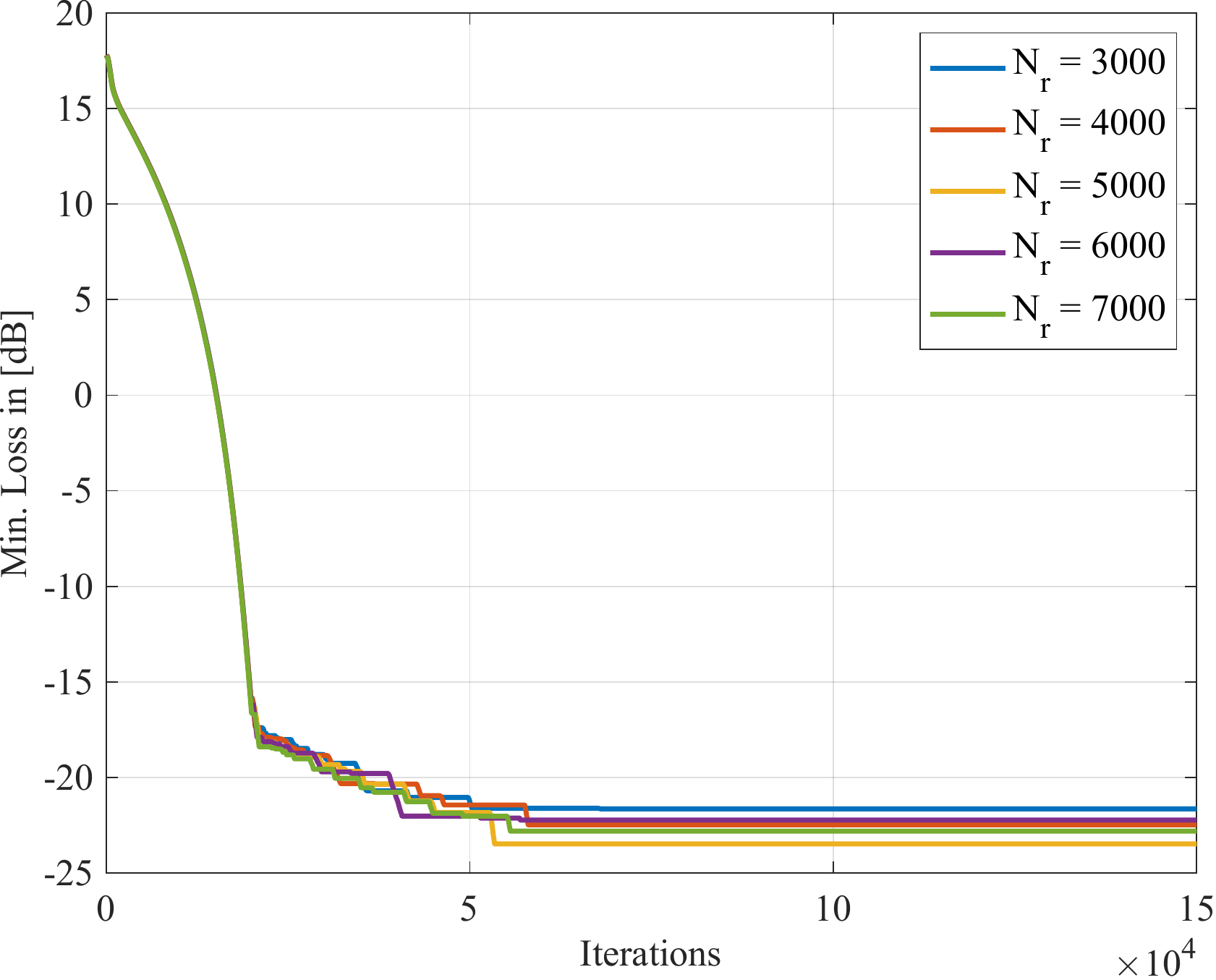}
        \caption*{(c)}
        \label{fig:subfigure3}
    \end{subfigure}
    \caption{ Impact of Network Architecture on Training Loss: (a) Layers, (b) Nodes, (c) Collocation points. }
    \label{fig:subfigures}
\end{figure}
\begin{figure}[ht]
    \centering
    \includegraphics[width=0.45\textwidth]{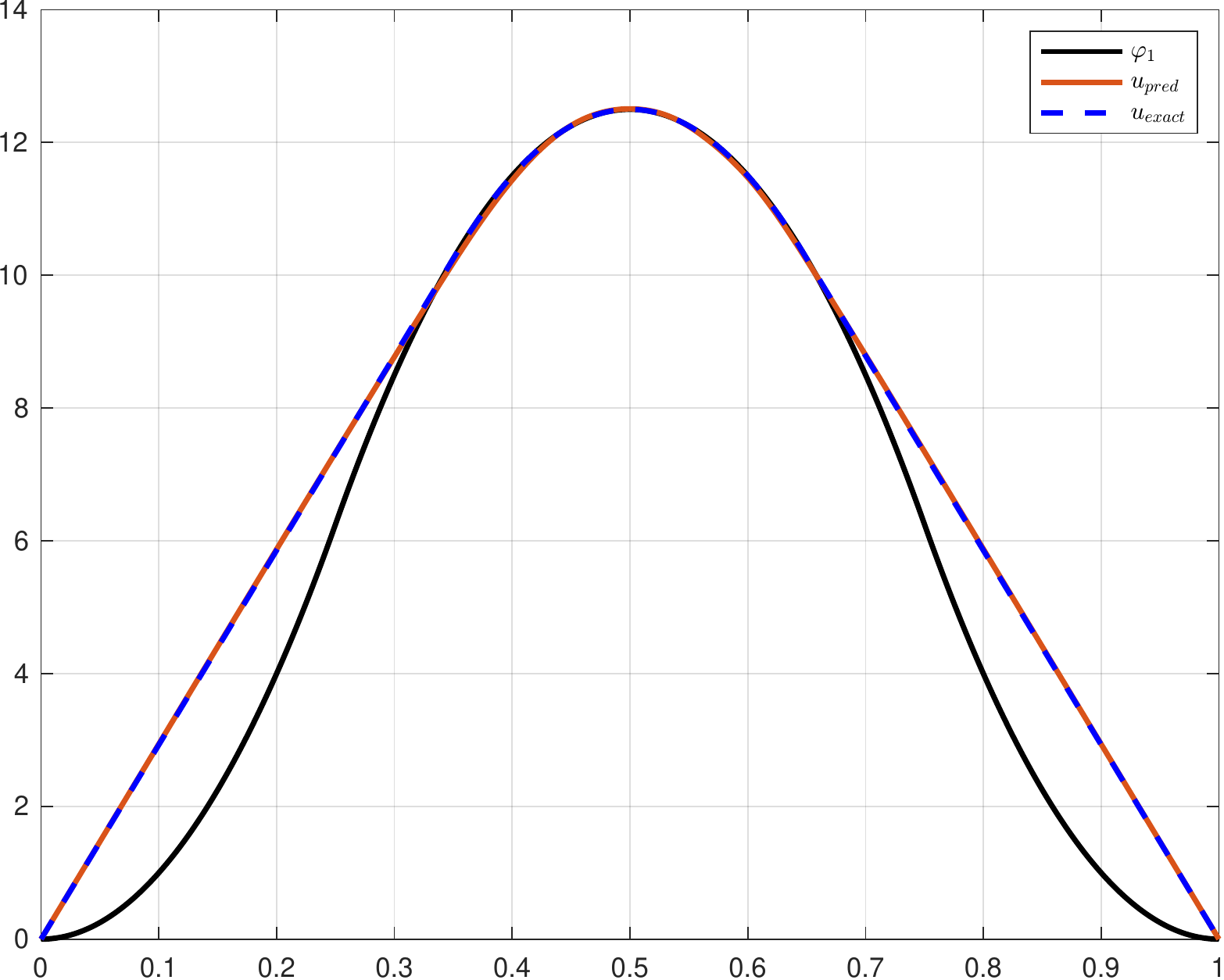}
     \includegraphics[width=0.45\textwidth]{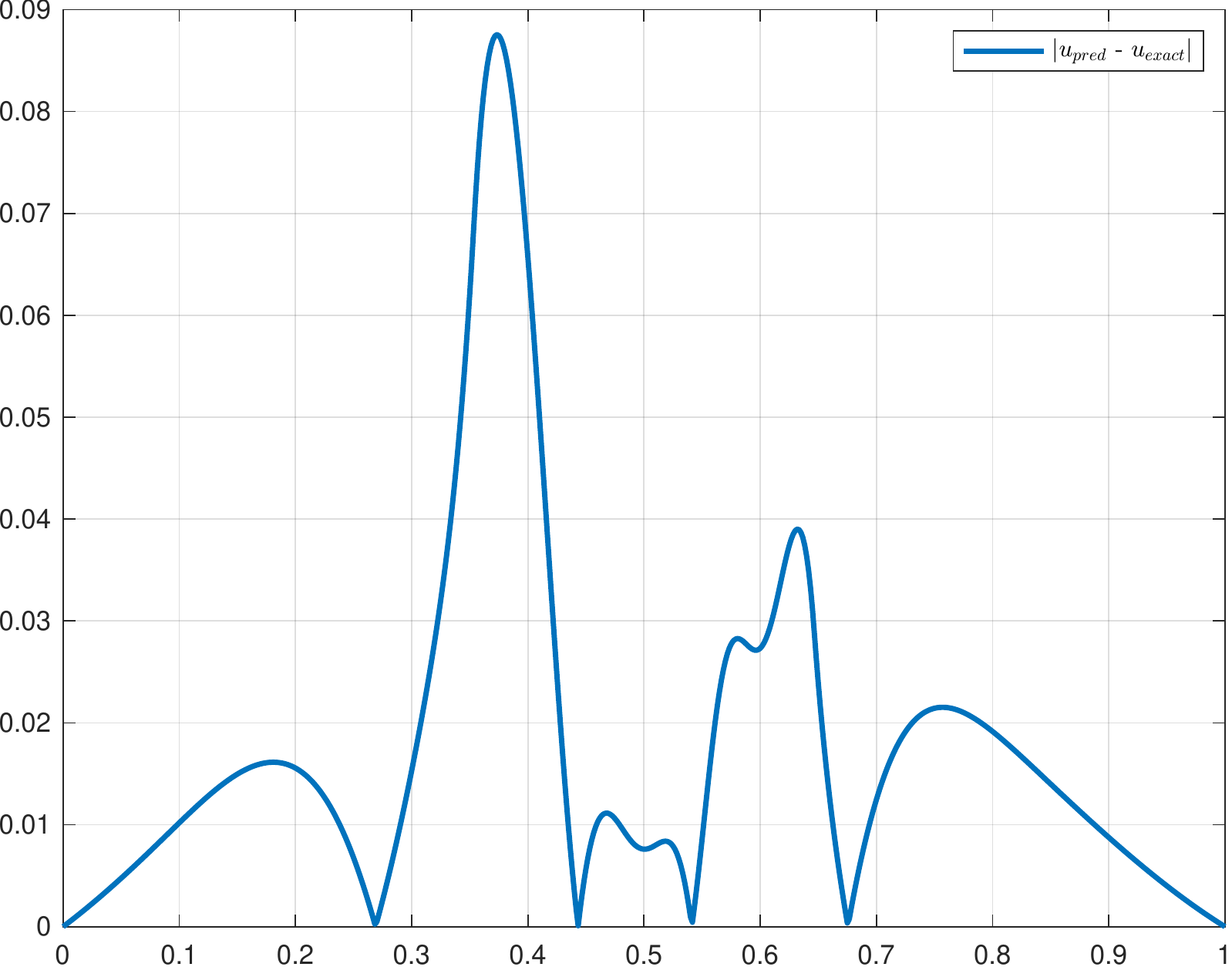}
    \caption{One-dimensional $\varphi_{1}$-obstacle Poisson's equation: (left) The predicted solution against the exact solution (4.4). (right)  A plot of the pointwise $L^{\infty}$-error estimation.  }
    \end{figure}
    
Subsequently, we can demonstrate that PINNs solution defined as in (4.5), also gives a good approximation to the solution of the Poisson equation (4.1) under another smooth obstacle $\varphi_2$ defined as follows
 \begin{equation}\varphi_{2}(x)= \begin{cases}
    10\sin(2\pi x)~~~~~&\text{for}~0\leq x\leq 0.25,\\
    5\cos(\pi(4x-1))+5~~~~~&\text{for}~0.25\leq x\leq 0.75,\\
    10\sin(2\pi(1-x))~~~~~&\text{for}~0.75\leq x\leq 1.
      \end{cases}\end{equation}
The corresponding exact solution such that $u\geq \varphi_{2}$ is
\begin{equation}u(x)=\begin{cases}
    10\sin(2\pi x)~~~~~&\text{for}~0\leq x\leq 0.25,\\
    10~~~~~&\text{for}~0.25\leq x\leq 0.75,\\
    10\sin(2\pi(1-x))~~~~~&\text{for}~0.75\leq x\leq 1.
      \end{cases}\end{equation}

\begin{figure}[ht]
    \centering 
    \includegraphics[width=0.45\textwidth]{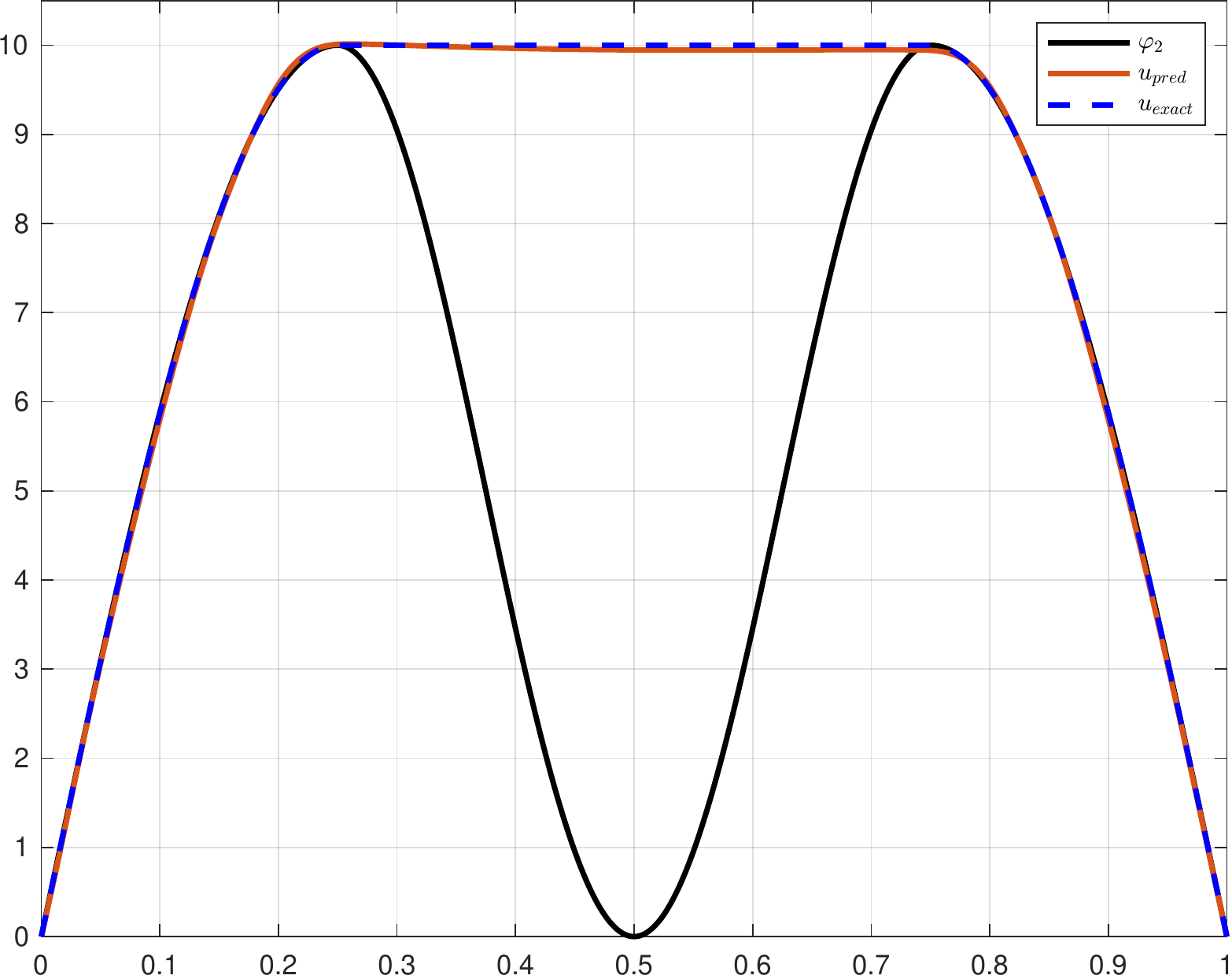}
     \includegraphics[width=0.45\textwidth]{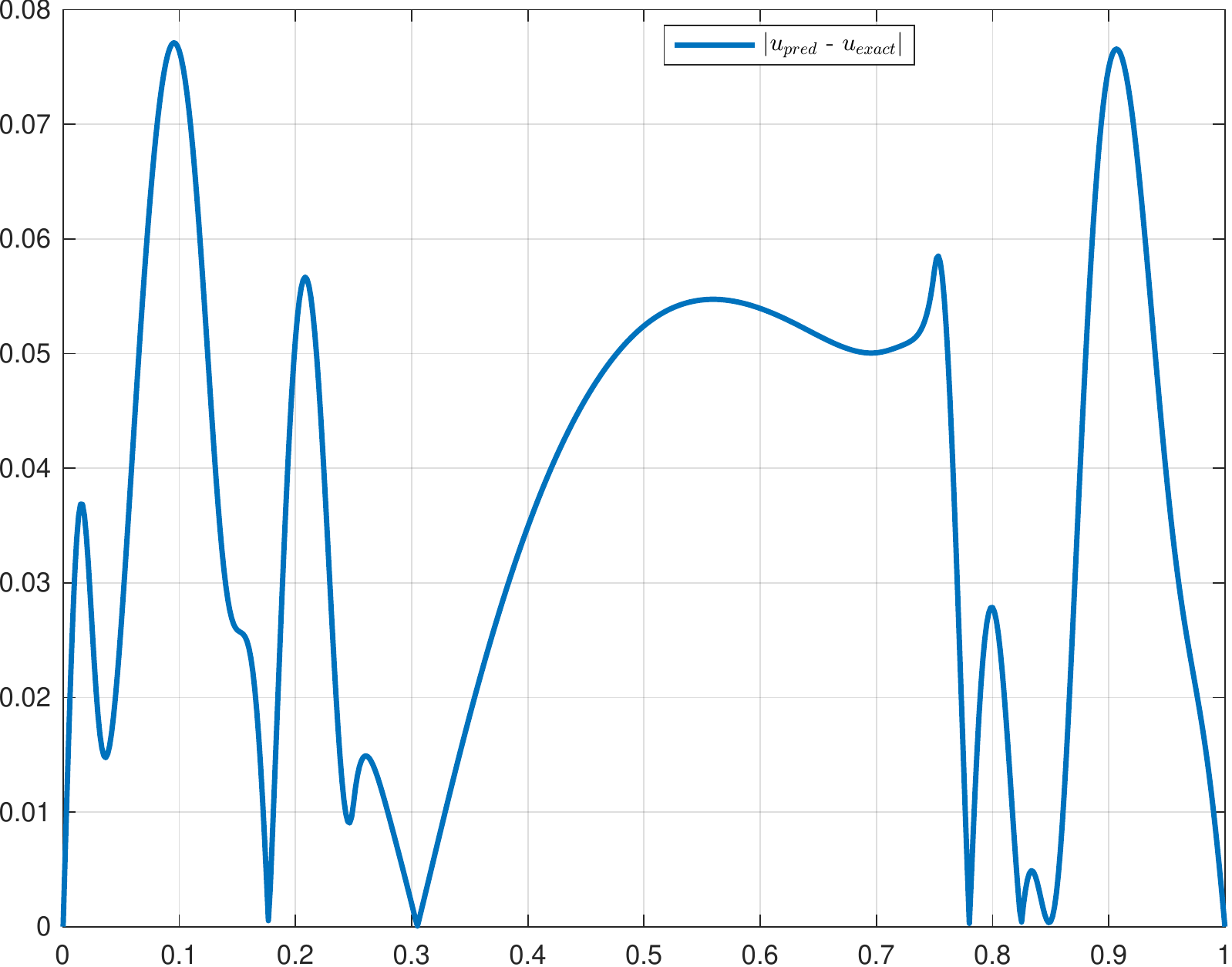}
    \caption{One-dimensional $\varphi_{2}$-obstacle Poisson's equation: (left) The predicted solution against the exact solution (4.7). (right)  A plot of the pointwise $L^{\infty}$-error estimation.}
\end{figure}
Figures 5 and 6 present visual comparisons with the exact solution $u$ for obstacles $\varphi_1$ and $\varphi_2$ and the Dirichlet boundary condition (4.2). As can be seen, the approximations are in good agreement with the exact solutions. These figures indicate that our framework can obtain accurate predictions with a relative $L^{\infty}$-error.

For the sake of completeness, we also considered an example with the same Poisson's equation with Dirichlet boundary  and an irregular obstacle $\varphi_3$ defined as follows
\begin{equation}\varphi_{3}(x)= \begin{cases}
    5x-0.75~~~~~&\text{for}~0.15\leq x < 0.2,\\
    1~~~~~&\text{for}~0.2\leq x\leq 0.8,\\
    -5x+4.25~~~~~&\text{for}~0.8 < x\leq 0.85,\\
    0~~~~~&\text{for}~\text{else}
      \end{cases}\end{equation}
\begin{figure}[ht]
    \centering 
    \includegraphics[width=0.45\textwidth]{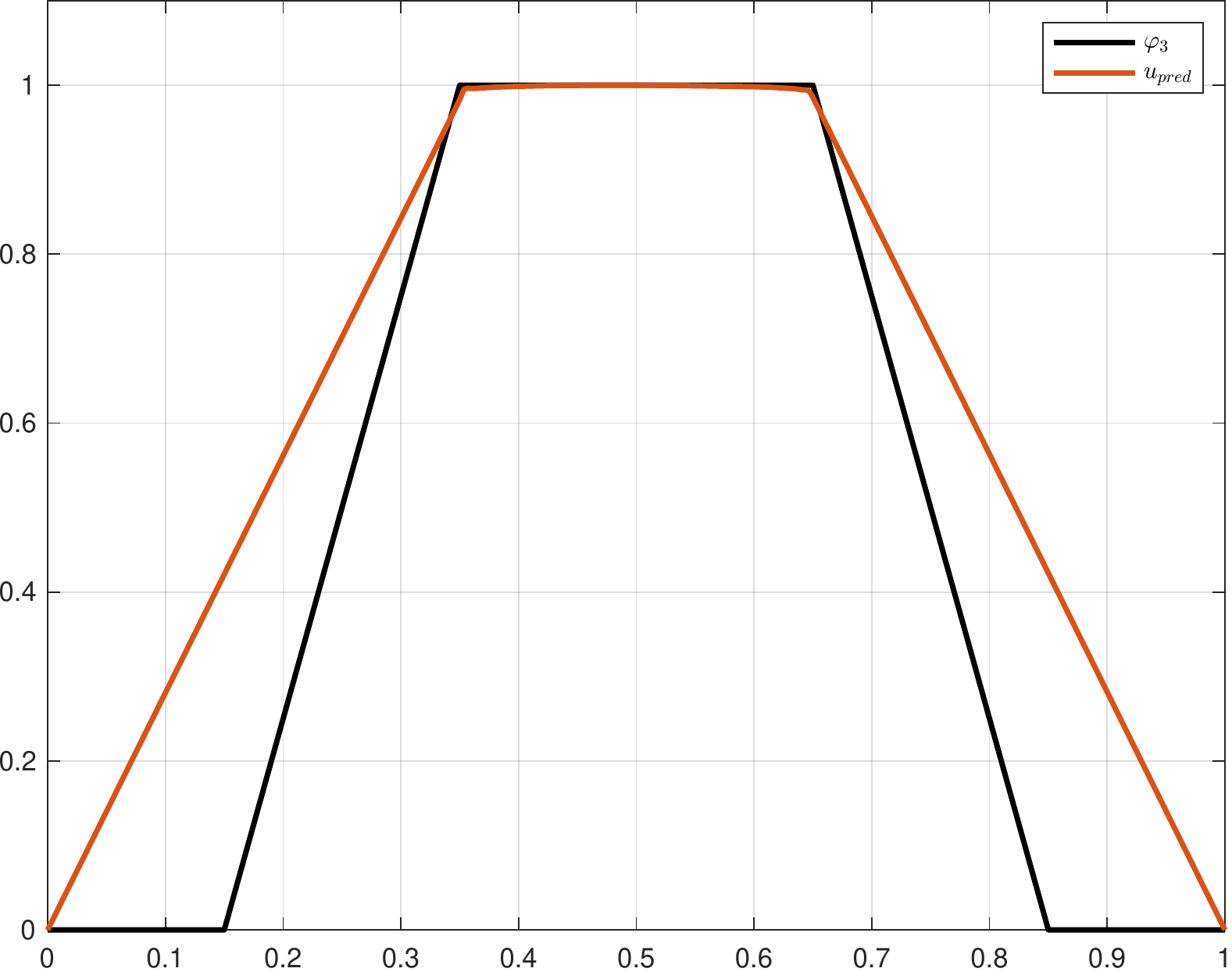}
    \caption{One-dimensional $\varphi_{3}$-obstacle Poisson's equation: The predicted PINNs solution.}
\end{figure}
This obstacle has been used and can be traced back to \cite{sch}, albeit with a typo in its definition. Therefore, applying Algorithm 1 to equation (4.1) with Dirichlet boundary (4.2) and under the irregular obstacle $\varphi_3$  gives a PINNs solution $\hat{u}(x,\theta)$ defined as in (4.5), which fulfills the boundary condition and it is above the obstacle $\varphi_3$ with some error in the rough corners of the contact area as it is shown in Figure 7. Therefore we cannot expect a very high accuracy similar to regular obstacles $\varphi_1$ and $\varphi_2$. This result is more accurate than the one given by the penalty method presented in \cite{sch}.

\subsection{One-dimensional nonlinear obstacle problem}
For our next case, we intend to demonstrate further that PINNs can also be applied to non-linear problems. Accordingly, we study a classical model of stretching a non-linear elastic membrane over a fixed obstacle defined as
\begin{equation}
   \displaystyle -\frac{\partial}{\partial x}\left( \frac{\frac{\partial u}{\partial x}}{\sqrt{(1+\left|\frac{\partial u}{\partial x}\right|^2)}}\right)=0,~~~\text{for}~x\in\Omega=(0,1),
   \end{equation}
subject to non-homogeneous Dirichlet boundary condition
\begin{equation}
   u(0)=5,~u(1)=10,
\end{equation}
and to an obstacle $\varphi_4$ which is defined by the following oscillatory function 
 \begin{equation}
  \varphi_{4}(x)= 10\sin^{2}{\pi(x+1)^2}
  \end{equation}
such that $u\geq \varphi_{4}$   on $\Omega=(0,1)$. Therefore, by using Algorithm 1 for equation (4.9) according to constraints (4.10) and (4.11), the PINNs solution
\begin{equation*}
 \hat{u}(x;\theta)  = 5(1-x)+ 10x + x(1-x)u(x;\theta) 
\end{equation*}
is identical to the obstacle on the contact set, and straight lines away from it as can be seen in Figure 8 which is similar to the results that appeared in \cite{l1,zo}.
\begin{figure}[ht]
    \centering 
    \includegraphics[width=0.45\textwidth]{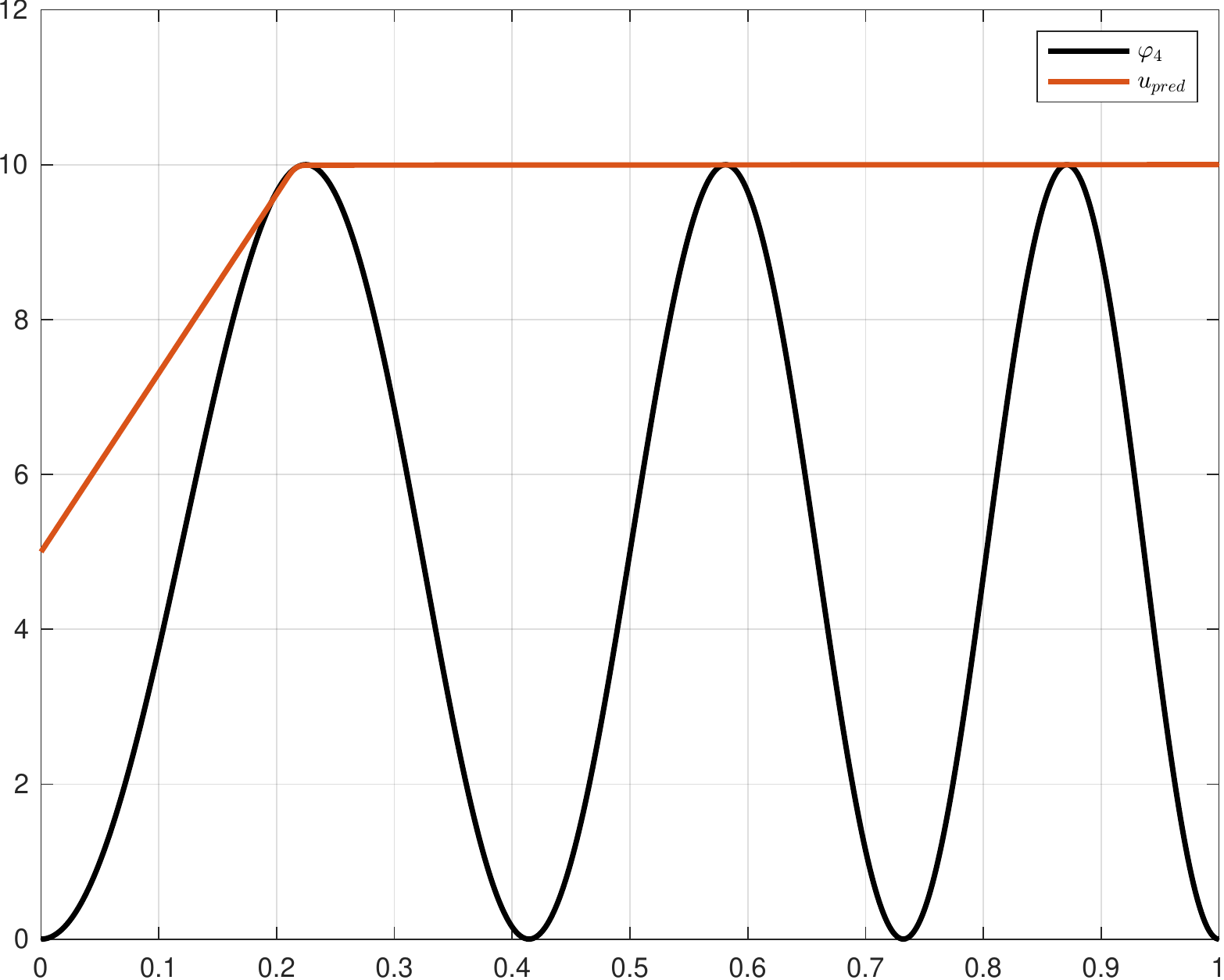}
    \caption{The PINNs solution of the nonlinear obstacle problem (4.9)-(4.11)}
\end{figure}
\subsection{Two-dimensional obstacle }
Our next numerical example is a two-dimensional (2D) problem defined on the domain $\Omega=[-2,2]\times[-2,2]$ such that
\begin{equation}
    -\frac{\partial^2 u}{\partial^2 x}-\frac{\partial^2 u}{\partial^2 y}=0,~\text{on}~\Omega,
\end{equation}
with the following obstacle 
\begin{equation}
   \varphi_{5}(x,y)= \begin{cases}
   \sqrt{1-x^{2}-y^{2}},&~~\text{for}~x^{2}+y^{2}\leq1,\\
   -1,&~~\text{otherwise}.
    \end{cases}
\end{equation}
This obstacle-related equation has been widely used by many authors to show the accuracy  of their proposed method \cite{l1,zo}. Since $\varphi_5$ is a radially-symmetric obstacle, the analytical solution of (4.12) subject to (4.13) is also radially-symmetric such that
\begin{equation}
   u(x,y)= \begin{cases}
   \sqrt{1-x^{2}-y^{2}},&~~~\text{for}~x^{2}+y^{2}\leq\beta,\\
   &\\
    \displaystyle-\beta^{2}\frac{\log\left(\frac{\sqrt{x^{2}+y^{2}}}{2}\right)}{\sqrt{1-\beta^{2}}}~~~&\text{otherwise},
    \end{cases}
\end{equation}
where $\beta= 0.6979651482..$. which satisfies $ \beta^{2}\left(1-\log\left(\frac{\beta}{2}\right)\right)=1.$
\begin{figure}[ht]
    \centering 
     \includegraphics[width=0.45\textwidth]{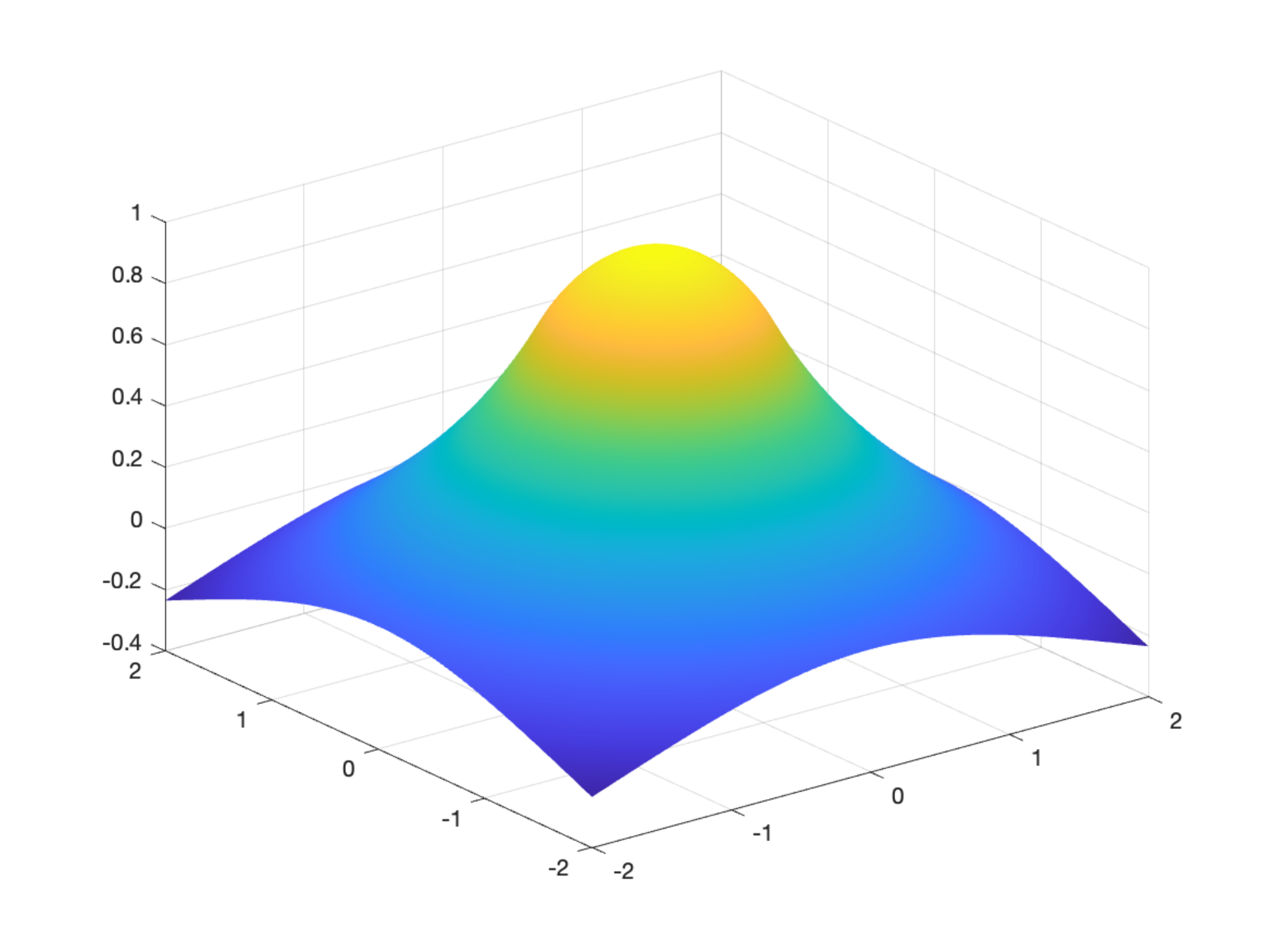}
      \includegraphics[width=0.45\textwidth]{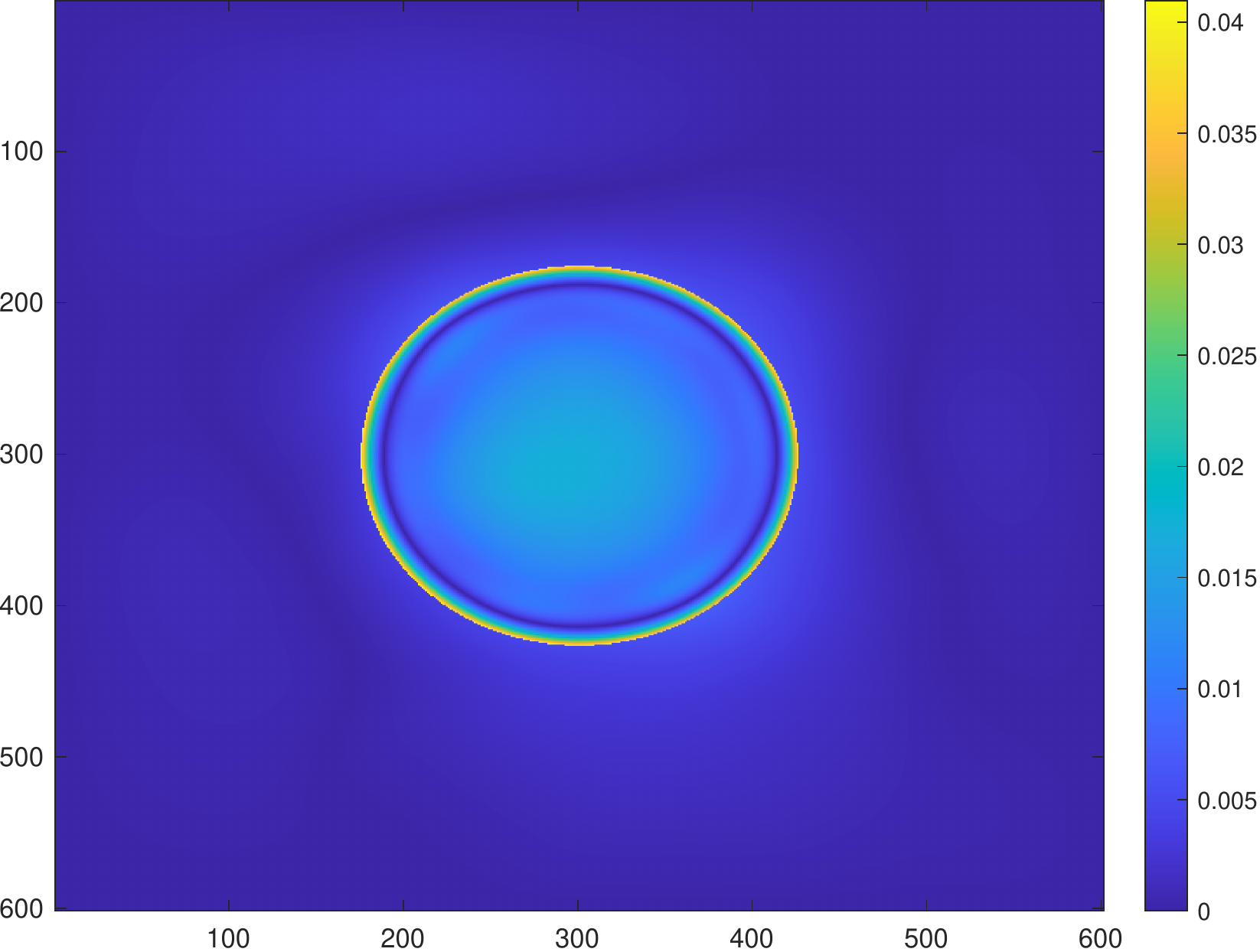}
    \caption{Radially symmetric 2D obstacle problem: (left) The predicted PINNs solution. (right) The relative pointwise $L^{\infty}$-error map.}
\end{figure}
The PINNs solution $\hat{u}(x,y;\theta)$ defined as
\begin{equation*}
    \hat{u}(x,y;\theta)= g(x,y)+(x-2)(x+2)(y-2)(y+2)u(x,y;\theta),
\end{equation*}
where 
\begin{equation*}
\begin{split}
g(x,y) = & -\beta^2 \left( \frac{\log\left(\frac{\sqrt{x^2 + 4}}{2}\right)}{\sqrt{1 - \beta^2}}  + \frac{\log\left(\frac{\sqrt{y^2 + 4}}{2}\right)}{\sqrt{1 - \beta^2}} \right)  + \beta^2 \left( \frac{\log\left(\frac{\sqrt{8}}{2}\right)}{\sqrt{1 - \beta^2}} \right),
\end{split}
\end{equation*}
and the $L^{\infty}$-pointwise difference with the exact solution are presented in Figure 9. It can be seen that the error is focused as peaks near the free boundary, where the function $u(x,y)$ is no longer $C^{2}$ (two times differentiable on $\Omega$) and is relatively small elsewhere.

For our last numerical example, we introduce a nonlinear non-harmonic equation of the form
\begin{equation*}
    \begin{cases}
        -\text{div}\left(|\nabla u|^{p-2}\nabla u\right)+1=0,&~~~~\text{in}~\Omega,\\
        u=0,&~~~~\text{on}~\partial\Omega,
    \end{cases}
\end{equation*}
which is known as the anisotropic p-Laplacian equation, which is related to many physical and biological models \cite{ham,ham2}. We test the PINNs framework for the $(p=4)$-Laplacian over the domain $\Omega=[0,2]\times[0,2]$ subject to the following irregular obstacle
\begin{equation*}
    \varphi_{6}(x,y)=\begin{cases}
       1,&~~~~\text{for}~0.5\leq x\leq 1.5,\\
       0,&~~~~\text{otherwise}.
    \end{cases}
\end{equation*}
As shown in \cite{l3}, the related exact solution is the following
\begin{equation*}
    u(x,y)=\begin{cases}
       \frac{3}{4}|x+7.75086|^{\frac{4}{3}}-11.50434,&~~~~\text{if}~x<0.5,\\
       1,&~~~~\text{if}~0.5\leq x\leq 1.5,\\
       \frac{3}{4}|-x+9.75086|^{\frac{4}{3}}-11.50434,&~~~~\text{if}~x>1.5.
    \end{cases}
\end{equation*}
\begin{figure}[ht]
    \centering 
     \includegraphics[width=0.45\textwidth]{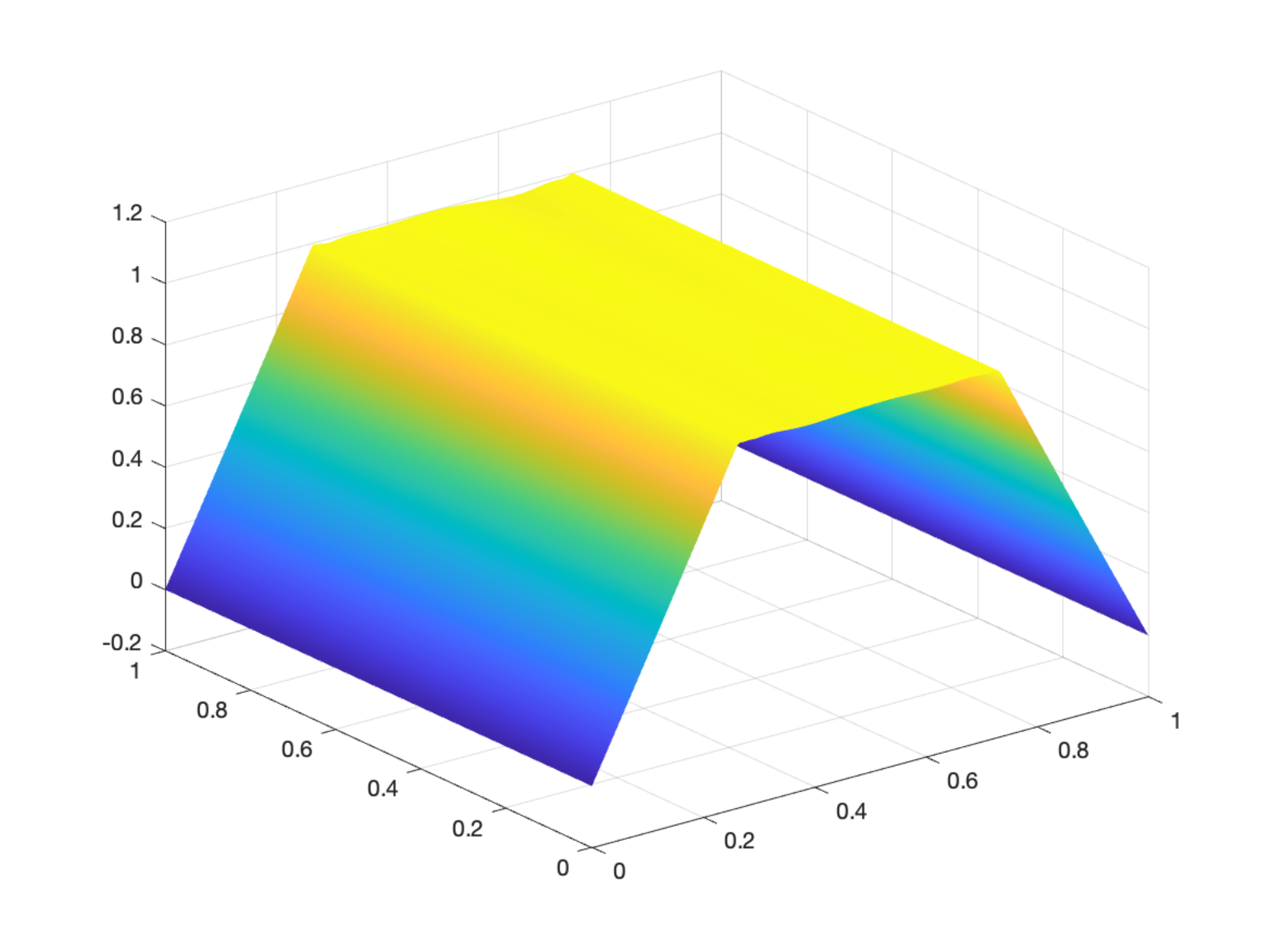}
      \includegraphics[width=0.45\textwidth]{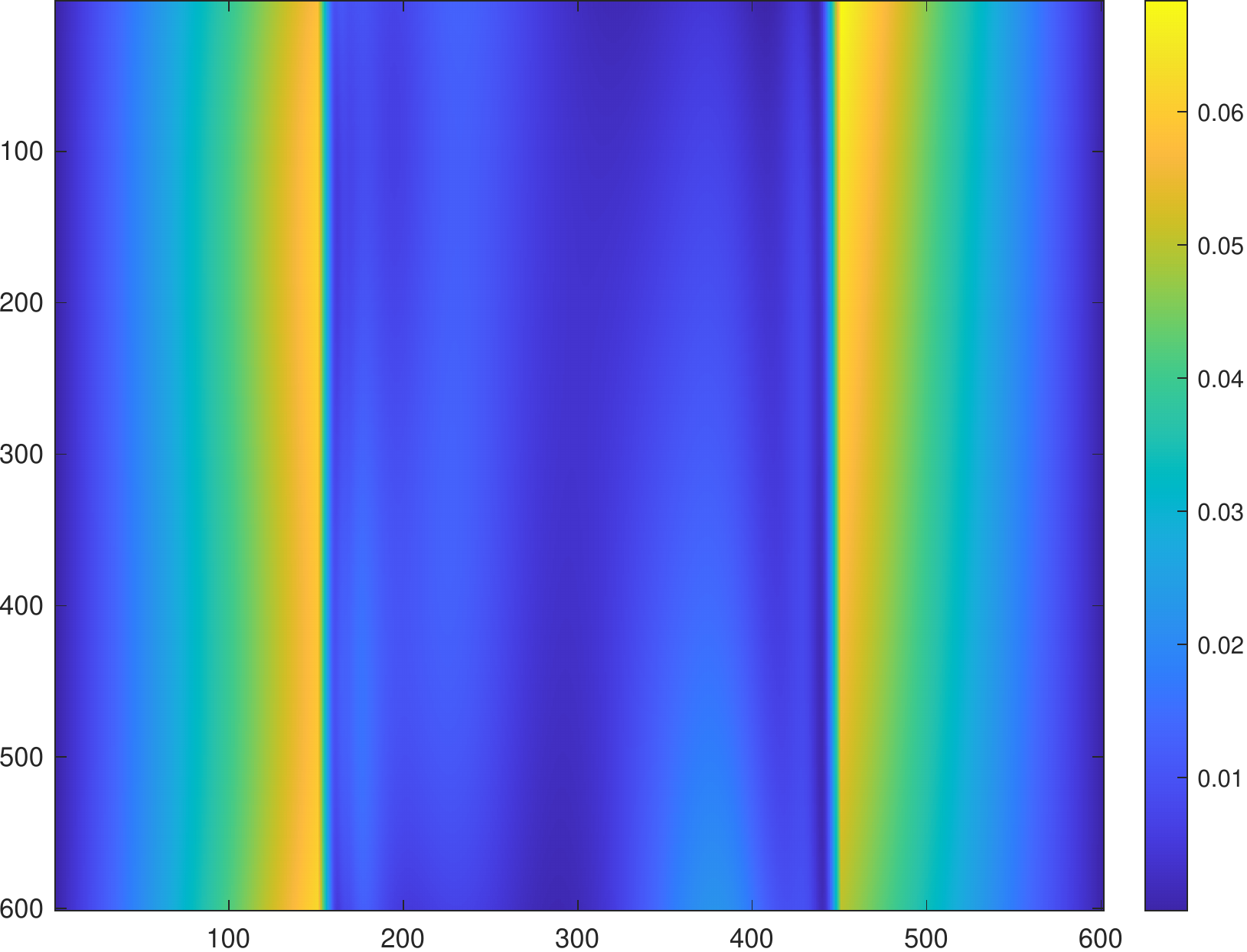}
    \caption{Quasi-linear 2D $p$-Laplacian obstacle problem: (left) The predicted PINNs solution. (right) The relative pointwise $L^{\infty}$-error map.}
\end{figure}
The PINNs solution $\hat{u}(x,y;\theta)$ is defined as 
\begin{equation*}
   \hat{u}(x,y;\theta)= u(0,y)\frac{2-x}{2} + u(2,y)\frac{x}{2} + x(2-x)u(x,y;\theta), 
\end{equation*}
and the pointwise $L^{\infty}$-error between exact and PINNs solution are presented in Figure 10. Despite the discontinuity of the obstacle which affects the regularity of the solution, the PINNs solution is smooth away from the obstacle and agrees well with the obstacle on its support. Also, it can be seen that the error is concentrated on the boundary of the obstacle, where the function $u(x,y)$ is discontinuous and is relatively small elsewhere.

\section{Conclusions and Outlook}
In summary, we present a PINNs framework with hard boundary constraints for modeling obstacle-related PDEs. The distinguishing feature of this deep learning method is the ability to train simultaneously the residual of the PINNs solution to be close to zero if the obstacle constraint is satisfied and to train the PINNs solution to be above the obstacle if else, which gives us the ability to predict the equilibrium position of the solution whose boundary is fixed and which is constrained to lie above the given obstacle. Furthermore, we have tested the performance of PINNs through a number of numerical cases with different formulations of the classical one- and two-dimensional PDEs with regular and irregular obstacles. As shown by the numerical experiments presented here, the resulting framework is general and flexible in the sense that it needs minimal implementation effort in order to be adapted to different kinds of obstacle problems. The PINNs framework doesn't require a high amount of labeled data pre-generated using high-fidelity simulation tools. It also does not depend on the discretization of the domain, which is typically the case in classical numerical approaches.

Despite the success exhibited by the proposed PINNs, it still has some limitations such as the cost of the computation is usually much larger than that of the methods mentioned in the introductory section due to the fact that a large number of optimization iterations are needed, but this can be improved by off-line training \cite{win,zhu}. We might also modify the NN architecture including NN width and depth, the activation function type, and interconnections between various hidden layers, particularly by cutting and adding certain connections. We can also automatically adjust these features of the neural network architecture by using meta-learning techniques \cite{fin,zop}. Another difficult question is how to tackle the problem of irregular obstacles with irregular geometry that may involve, mushy regions, discontinuities, or sharp cusps which we are trying to solve in our future work.  One possible way to tackle this problem is by using the Residual Adaptive Refinement RAR-PINNs \cite{rar}. This technique involves adaptively refining the neural network's focus on areas where it currently performs poorly, based on the residual errors in its predictions. On the other hand, we will also investigate some interesting related problems, particularly the double obstacle problems where the function is constrained to lie above one obstacle function and below another, level surfaces, and free boundary problems.

\section*{Funding}
This work was partially supported by DAAD grants 57417688 and 57512510 and the Carnegie Corporation of New York grant (provided through the AIMS Research and Innovation Centre). 
\section*{Data availability statement} The data sets generated or analyzed
during the current study are available from the corresponding
author on reasonable request.
\section*{Conflict of interest} The authors declare no conflict of interest.

\end{document}